\documentclass[11pt,a4paper]{article}
\usepackage[hyperref]{emnlp-ijcnlp-2019}
\usepackage{times}
\usepackage{latexsym}
\usepackage{fancyhdr,graphicx,amsmath,amssymb}

\usepackage[utf8x]{inputenc}
\usepackage[thaicjk,english]{babel}
\addto\extrasthaicjk{\fontencoding{C90}\selectfont}

\makeatletter
\@namedef{opt@inputenc.sty}{utf8}
\makeatother
\usepackage{CJKutf8}

\usepackage{multirow}
\usepackage{array}
\usepackage{url}

\aclfinalcopy 

\title{Zero-shot Cross-lingual Dialogue Systems with  \\ Transferable Latent Variables}

\author{Zihan Liu, Jamin Shin, Yan Xu, Genta Indra Winata, \\ \textbf{Peng Xu, Andrea Madotto, Pascale Fung} \\
Center for Artificial Intelligence Research (CAiRE)\\
Department of Electronic and Computer Engineering\\
The Hong Kong University of Science and Technology, Clear Water Bay, Hong Kong\\
\texttt{\{zliucr,jay.shin,yxucb,giwinata,pxuab\}@connect.ust.hk},\\ \texttt{pascale@ece.ust.hk}}

\date{}

\begin{document}
\maketitle
\begin{abstract}
Despite the surging demands for multilingual task-oriented dialog systems (e.g., Alexa, Google Home), there has been less research done in multilingual or cross-lingual scenarios.  Hence, we propose a zero-shot adaptation of task-oriented dialogue system to low-resource languages. To tackle this challenge, we first use a set of very few parallel word pairs to refine the aligned cross-lingual word-level representations. We then employ a latent variable model to cope with the variance of similar sentences across different languages, which is induced by imperfect cross-lingual alignments and inherent differences in languages. Finally, the experimental results show that even though we utilize much less external resources, our model achieves better adaptation performance for natural language understanding task (i.e., the intent detection and slot filling) compared to the current state-of-the-art model in the zero-shot scenario.
\end{abstract}

\section{Introduction}

Task-oriented dialogue systems have been widely adopted in the industry (e.g., Amazon Alexa, Google Home, Apple Siri, Microsoft Cortana) as a virtual agent to tend to the needs of the users. However, these agents have mostly been trained with the monolingual dataset that is often expensive to build or acquire. In order to cope with the scarcity of low-resource language dialogue data, we are motivated to look into cross-lingual dialogue systems which can adapt with very little or no training data in the target language.

\begin{figure}[t!]
\centering
\includegraphics[scale=0.11]{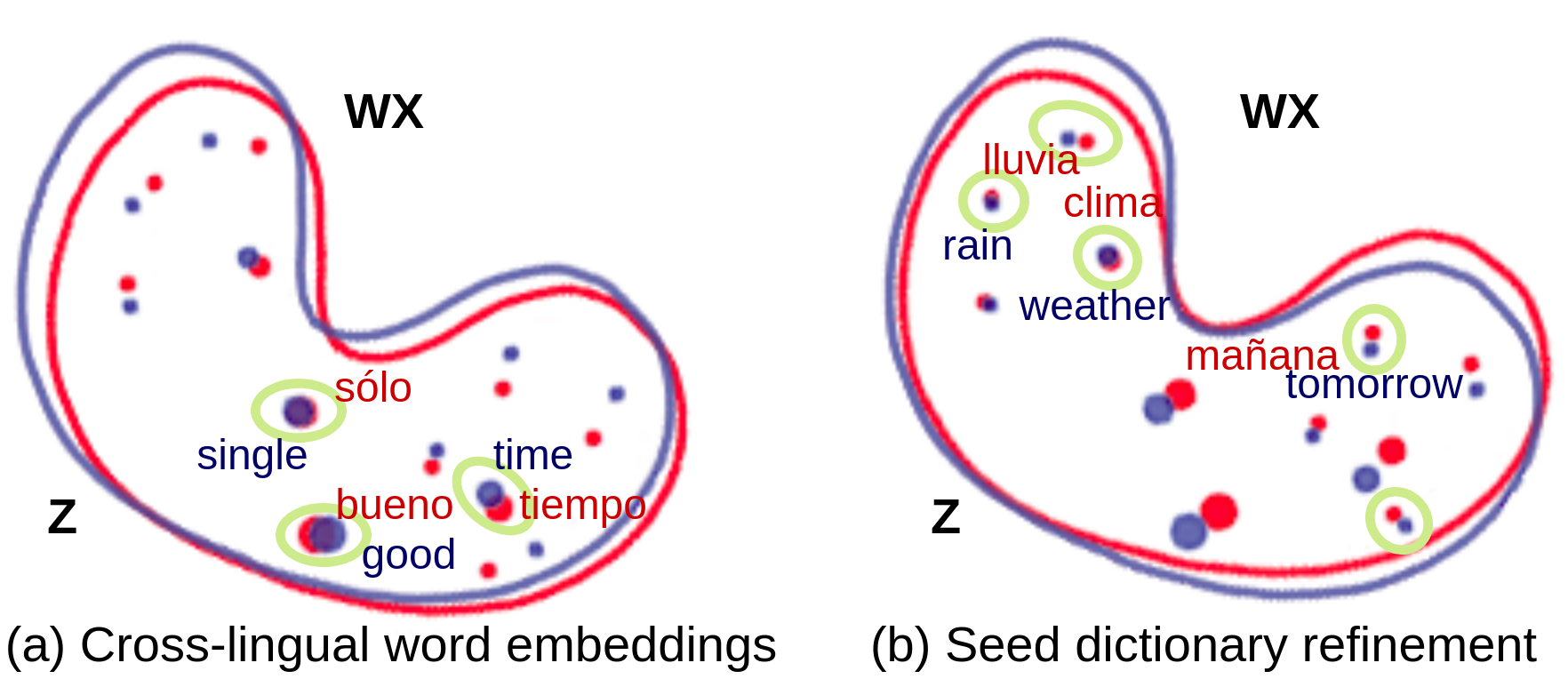}
\caption{(a) The aligned word embeddings. (b) Refined mapping $\mathbf{W}$ with seed dictionary; green circle represents close words in different languages.}
\label{fig:emb}
\end{figure}

This task of zero-shot adaptation of dialogue systems to different languages is relatively new and has not been explored thoroughly enough yet. The main approach of previous work~\cite{upadhyay2018almost,chen2018xl,schuster2019cross} in this task is using aligned cross-lingual word embeddings between source and target languages.  
However, this method suffers from imperfect alignments between the source and target language embeddings. This can be attributed not only to the noise in aligning two different embeddings, but also to the inherent discrepancies in different languages such as Thai and English which come from entirely different roots.
To address such variance in the alignment, we turn to probabilistic modeling with latent variables as it has been successfully used in several recent task-oriented dialogue systems~\cite{wen2017latent,zhao2017learning,zhao2018unsupervised,zhao2018zero, le2018variational}. 

However, we notice that naively using latent variables does not help the model improve much in slot filling and intent prediction. We hypothesize that the variance of the cross-lingual word embeddings is too large for the model to learn any meaningful latent variables. Hence, we propose to first refine the cross-lingual embeddings with $\sim$10 seed word-pairs related to the dialogue domains. We then add Gaussian noise~\cite{Zheng_2016_CVPR} to further compensate the imperfect alignment of cross-lingual embeddings.

As a result, a combination of these methods allows us to build a transferable latent variable model that learns the distribution of training language inputs that is invariant to noise in the cross-lingual word embeddings. This enables our model to capture the variance of semantically similar sentences across different languages, and achieve state-of-the-art results in zero-shot adaptation of English to Spanish and Thai for the natural language understanding task (i.e., the intent prediction and slot filling) on the dataset proposed by \citet{schuster2019cross}, even though we use much less external resources (i.e., $\sim$10 seed word-pairs) while others utilize a large amount of bilingual corpus. We further visualize the learned latent variables to confirm that same-meaning words and sentences have similar distributions.

\begin{figure}[t!]
\centering
\includegraphics[scale=0.63]{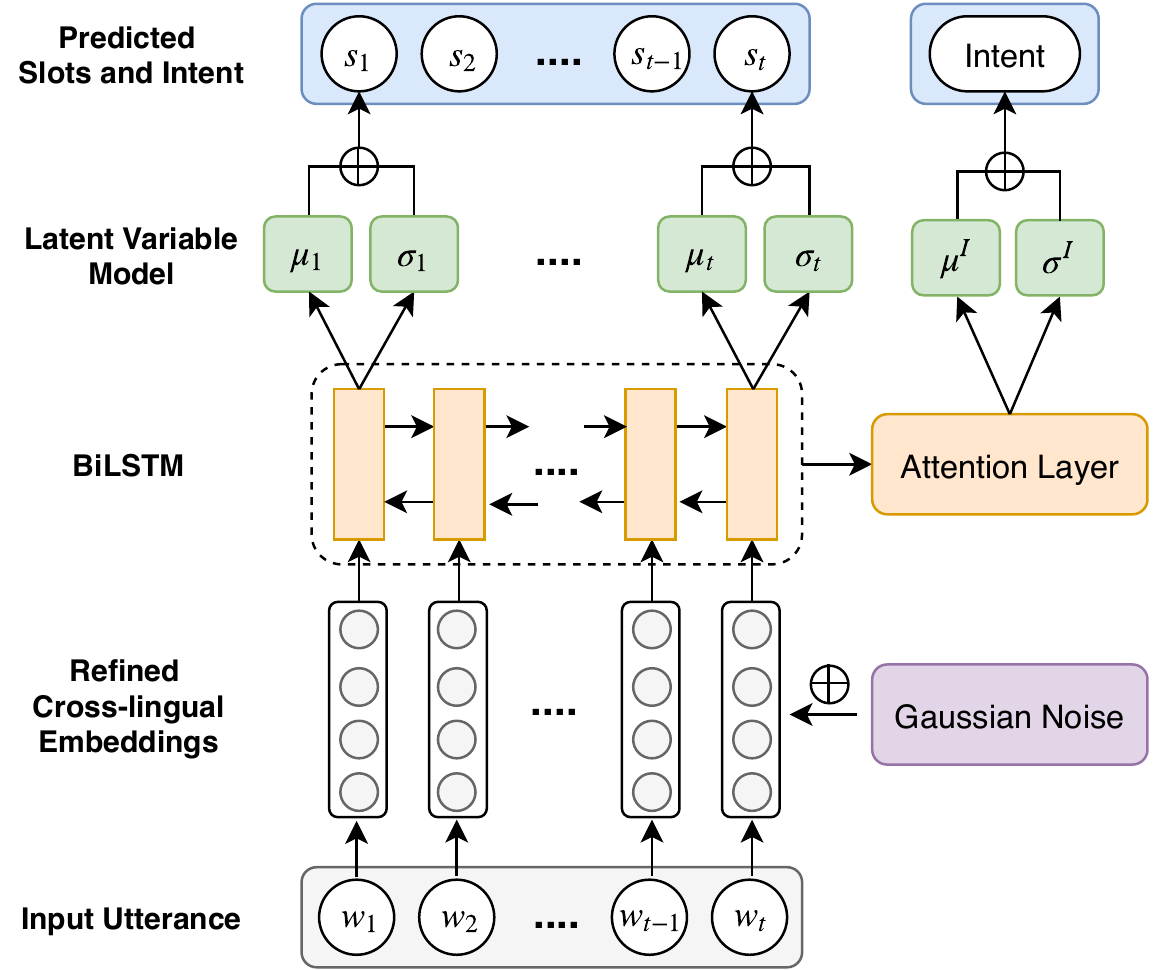}
\caption{The structure of our proposed model.}
\label{fig:model}
\end{figure}

\section{Related Work}
Cross-lingual transfer learning which acts as one of the low-resource topics~\cite{gu2018universal,lee2019team,liu2019incorporating,xu2018emo2vec} has attracted more and more people recently, followed by the rapid development of cross-lingual word embeddings. \citet{artetxe2017learning} proposed a self-learning framework and utilized a small size of word dictionary to learn the mapping between source and target word embeddings. \citet{conneau2017word} leveraged adversarial training to learn a linear mapping from a source to a target space without using parallel data. \citet{joulin2018loss} utilized Relaxed CSLS loss to optimize this mapping problem. \citet{winata2019learning} introduced a method to leverage cross-lingual meta-representations for code-switching named entity recognition by combining multiple monolingual word embeddings.

\citet{chen2018xl} proposed a teacher-student framework leveraging bilingual data for cross-lingual transfer learning in dialogue state tracking. \citet{upadhyay2018almost} leveraged joint training and cross-lingual embeddings to do zero-shot and almost zero-shot transfer learning in intent prediction and slot filling. Finally, \citet{schuster2019cross} utilizes Multilingual CoVe embeddings obtained from training Machine Translation systems as in~\cite{mccann2017learned}. The main difference of our work with previous work is that our model does not leverage any external bilingual data other than 11 word pairs for embeddings refinement.

\section{Methodology}
Our model consists of a refined cross-lingual embedding layer followed by a BiLSTM~\cite{hochreiter1997long} which parameterizes the Latent Variable Model, as illustrated in Figure~\ref{fig:model}. We jointly train our model to predict both slots and user intents. We denote $\mathbf{w} = [w_1, \dots, w_T]$ as the input words and $\mathbf{e} = [e_1, \dots, e_T]$ as the word embeddings of $\mathbf{w}$. The slot at time-step $t$ is $s_t$, while the intent for each sentence $\mathbf{w}$ is denoted as $I$. Note that only matrices are bold-faced.

\subsection{Cross-lingual Embeddings Refinement}
To further refine the cross-lingual alignments to our task, we draw from the hypothesis that domain-related words are more important than others. Hence, as shown in Figure \ref{fig:emb}, we propose to refine the cross-lingual word embeddings~\cite{joulin2018loss}\footnote{The embeddings are available in https://fasttext.cc} using very few parallel word pairs, which is obtained by selecting 11 English words related to dialogue domains (weather, alarm, and reminder) and translate them using bilingual lexicons. We refine the embeddings by leveraging the framework proposed  in~\citet{artetxe2017learning}.

Let $\mathbf{X}$ and $\mathbf{Z}$ be the aligned cross-lingual word embeddings between two languages. $X_{i*}$ and $Z_{j*}$ are the embeddings for the $i^{th}$ source word and $j^{th}$ target word. We denote a binary dictionary matrix $\mathbf{D}$: $D_{ij}=1$ if the $i^{th}$ source language word is aligned with the $j^{th}$ target language word and $ D_{ij}=0 $ otherwise. The goal is to find the optimal mapping matrix $\mathbf{W}^{*}$ by minimizing:
\begin{align}
    \mathbf{W}^{*} = \arg\min_\mathbf{W} \sum_{i,j} D_{ij} ||X_{i*} \mathbf{W} - Z_{j*}||^2. 
    \label{eq:1}
\end{align}
Following \citet{artetxe2016learning}, with orthogonal constraints, mean centering, and length normalization, we can maximize the following instead:
\begin{align}
    \mathbf{W}^{*} = \arg\max_{\mathbf{W}} \text{Tr}(\mathbf{X}\mathbf{W} \mathbf{Z}^\text{T} \mathbf{D}^\text{T}).
    \label{eq:2}
\end{align}
We iteratively optimize Equation~\ref{eq:2} until distances between domain-related seed words are closer than a certain threshold after refinement. Figure~\ref{fig:emb} illustrates better alignment for domain-related words after refinement.

\begin{table*}[ht!]
\centering
\begin{tabular}{l|cc|cc|cccc}
\hline
\multicolumn{1}{c|}{}     & \multicolumn{4}{c|}{\textbf{Spanish}}  & \multicolumn{4}{c}{\textbf{Thai}}      \\ \hline
\multicolumn{1}{c|}{\multirow{2}{*}{Model}} & \multicolumn{2}{c|}{\textbf{Intent acc.}} & \multicolumn{2}{c|}{\textbf{Slot F1}} & \multicolumn{2}{c|}{\textbf{Intent acc.}}   & \multicolumn{2}{c}{\textbf{Slot F1}} \\ \cline{2-9} 
\multicolumn{1}{c|}{}                       & \textbf{LVM}         & \textbf{CRF}       & \textbf{LVM}       & \textbf{CRF}     & \textbf{LVM}            & \multicolumn{1}{c|}{\textbf{CRF}}   & \textbf{LVM}  & \textbf{CRF} \\ \hline
Vanilla BiLSTM    & 46.36    & 44.13   & 15.64    & 11.32    & 35.12  & \multicolumn{1}{c|}{33.57} & 5.82                  & 5.24         \\
\quad+ \textit{noise (N)}   & 72.97  & 66.95  & 46.56  & 20.27  & 40.37          & \multicolumn{1}{c|}{37.53} & 10.66    & 6.51         \\
\quad+ \textit{refinement (R)}  & 87.69    & 88.23    & 61.63  & 42.62            & 59.40   & \multicolumn{1}{c|}{59.28} & 21.84  & 16.53        \\
\quad+ \textit{noise \& refinement}  & 89.21  & 88.79 & 64.04  & 43.98            & 70.81 & \multicolumn{1}{c|}{64.48} & 29.54  & 17.46         \\
\quad+ \textit{N \& R \& delexicalization}      & \textbf{90.20}       & 89.98              & \textbf{65.79}     & 47.70            & \textbf{73.43} & \multicolumn{1}{c|}{69.62} & \textbf{32.24}        & 23.11  \\ \hline 
Zero-shot SLU $^\ddagger$                              & \multicolumn{2}{c|}{46.64}                & \multicolumn{2}{c|}{15.41}            & \multicolumn{2}{c|}{35.64}                  & \multicolumn{2}{c}{12.11}            \\
Multi. CoVe w/ auto                            & \multicolumn{2}{c|}{53.89}                & \multicolumn{2}{c|}{19.25}            & \multicolumn{2}{c|}{70.70}                  & \multicolumn{2}{c}{35.62}           \\ \hline\hline
Translate Train $^\dagger$        & \multicolumn{2}{c|}{\textit{85.39}}   & \multicolumn{2}{c|}{\textit{72.87}}  & \multicolumn{2}{c|}{\textit{95.85}} & \multicolumn{2}{c}{\textit{55.43}} \\
\hline
\end{tabular}
\caption{Results on different models including baseline models, where N refers to the Gaussian noise injection and R refers to the cross-lingual embeddings refinement.
$^\ddagger$ We implemented \citet{upadhyay2018almost} model and evaluated with our test set.  $^\dagger$ \citet{schuster2019cross} translated English data to Spanish and Thai with a trained supervised machine translation system and it is considered as our \textit{upper bound} result.}
\label{tab:result}
\end{table*}
\subsection{Gaussian Noise Injection}
To cope with the noise in alignments, we inject Gaussian noise to English embeddings, so the trained model will be more robust to variance. This is a regularization method to improve the generalization ability to the unseen inputs in different languages, particularly languages from different roots such as Thai and Spanish. The final embeddings are $\mathbf{e}^* = [e_1 + N_1, \dots, e_T + N_T]$, where $\mathbf{N} \sim \mathcal{N}(0, 0.1 \mathbf{I})$.

\subsection{Latent Variable Model (LVM)}
\label{lvm}
Given a near-perfect cross-lingual embedding, there is still noise caused by the inherent discrepancies between source and target languages. This noise amplifies when combined with imperfect alignment, and makes point estimation vulnerable to the small, but not negligible differences across languages.
Instead, using latent variables will allow us to model the distribution that captures the variance of semantically similar sentences across different languages. The whole training process is defined as follows:
\begin{gather}
    [h_1...h_t...h_T] = \text{BiLSTM}(\mathbf{e}^*), \\
    m_t = h_t w_a,~ a_t = \frac{exp(m_t)}{\sum_{j=1}^T exp(m_j)},~ v = \sum_{t=1}^T a_t h_t, \\
    \left[ \begin{array}{c} { \mu^S_t } \\ { \log \left (\sigma^S_t) ^ { 2 } \right. } \end{array} \right] = \mathbf{W}^S_r h_t, \left[ \begin{array} { c } { \mu^I } \\ { \log \left (\sigma^I) ^ { 2 } \right. } \end{array} \right] = \mathbf{W}^I_{r} v, \\
    z^S_t \sim q^S_t (z | h_t), ~~ z^I \sim q^I (z | v), \\
    p^S_t (s_t | z^S_t) = \text{Softmax}(\mathbf{W}^S_g z^S_t), \\
    p^I (I | z^I) = \text{Softmax}(\mathbf{W}^I_g z^I),
\end{gather}
where attention vector ($v$) is obtained by following~\citet{felbo2017using} and $w_a$ is the weight matrix for the attention layer, $\mathbf{W}_{\{r,g\}}^{\{S,I\}}$ are trainable parameters, superscripts S and I refer to slot prediction and intent detection respectively, subscript ``r'' refers to ``recognition'' for obtaining the mean and variance vectors while subscript ``g'' refers to ``generation'' for predicting the slots and intents, and $q^S_t \sim \mathcal{N}(\mu^S_t, (\sigma^S_t)^2\mathbf{I})$ and $q^I \sim \mathcal{N}(\mu^I, (\sigma^I)^2\mathbf{I})$ are the posterior approximations which we sample our latent vectors $z^S_t$ and $z^I$ from. Finally, $p^S_t$ and $p^I$ are the predictions for the slot of the \textit{t-th} token and the intent of the utterance respectively. The objective functions for slot filling and intent prediction are:
\begin{gather}
    \mathcal{L}^I = \mathbb{E}_{z^I} [\log p^I (I|z^I)], \\
    \mathcal{L}^S_t = \mathbb{E}_{z^S_t} [\log p^S_t (s_t|z^S_t)], \\
    \mathcal{L}^S = \sum_{t=1}^T L_t^S,
\end{gather}
hence, the final objective function to minimize is,
\begin{equation}
    \mathcal{L} = \mathcal{L}^S + \mathcal{L}^I.
\end{equation}
The model prediction is not deterministic since the latent variables $z_t^S$ and $z^I$ are sampled from the Gaussian distributions. Therefore, in the inference time, we use the true mean $\mu_t^S$ and $\mu^I$ to replace $z_t^S$ and $z^I$ respectively to make the prediction deterministic.

\section{Experiments}

\subsection{Dataset}
We conduct our experiments under the zero-shot scenario of multilingual task-oriented dialogue dataset presented by \citet{schuster2019cross}. Our model is trained only with the English data and then do a zero-shot test on Spanish and Thai test set. We delexicalize words by replacing the tokens which represent numbers, time (such as am, pm), and duration (such as 30min) with special tokens \texttt{<number>}, \texttt{<time>}, and \texttt{<last>} respectively.

\subsection{Training Details}
In the training procedure, we freeze the word embeddings of the primary language, and then replace them with the corresponding aligned word embeddings of the unseen languages for a zero-shot test. We use bi-directional LSTM model with hidden dimension size of 250, and the latent variable model with both mean and variance in the size of 100. Gaussian noise with zero mean and variance of 0.1 is injected dynamically in different iterations. We use the accuracy to evaluate the performance of intent prediction and the standard BIO structure to calculate the F1 score for evaluating the performance of slot filling. In the zero-shot cross-lingual adaptation, we simply replace the training language (i.e., English) word embeddings with the cross-lingual target language (i.e., Spanish or Thai) word embeddings. Note that we never use any target language evaluation data to select the model for zero-shot cross-lingual adaptation, instead, we utilize the English validation set and early stop strategy according to the slot F1 score. 

\subsection{Word Pairs}
We choose the number of word pairs based on the vocabulary size of the corpus. Intuitively, the larger the vocabulary size is, the more words we need to align across languages, and the more word pairs we need to achieve good performance. We select 11 domain-related words which frequently exist in the English training set. The number of words we select is around 0.25\% of the vocabulary size for the English training set. The concrete information of the 11 word pairs is as follows:

The English seed words we selected are \textit{weather, forecast, temperature, rain, hot, cold, remind, forget, alarm, cancel, tomorrow}, which are related to the three dialogue domains (weather, alarm, and reminder). We translate them by leveraging bilingual dictionaries\footnote{https://github.com/facebookresearch/MUSE}. The corresponding translations in Spanish and Thai are \textit{clima, pronóstico, temperatura, lluvia, caliente, frío, recordar, olvidar, alarma, cancelar, mañana} and 
\foreignlanguage{thaicjk}{\textit{อากาศ, พยากรณ์, อุณหภูมิ, ฝน, ร้อน, หนาว, เตือน, ลืม, เตือน, ยกเลิก, พรุ่ง }} respectively.

\begin{table}[!t]
\begin{tabular}{l|cc|cc}
\hline
\multicolumn{1}{c|}{\multirow{2}{*}{Model}} & \multicolumn{2}{c|}{Spanish}    & \multicolumn{2}{c}{Thai}        \\ \cline{2-5} 
\multicolumn{1}{c|}{}                       & Intent         & Slot           & Intent         & Slot           \\ \hline
Our model    & \textbf{90.20} & \textbf{65.79} & \textbf{73.43} & \textbf{32.24} \\
\textit{- LVM}  & 85.85   & 61.86   & 66.01    & 25.22          \\
\textit{- LVM + MLP}  & 86.02   & 62.34  & 66.56   & 28.35   \\
\hline
\end{tabular}
\caption{Ablation Study on LVM models. \textit{\{- LVM\}} means removing LVM and \textit{\{- LVM + MLP\}} means replacing LVM with a Multi-Layer Perceptron which has the same size as the LVM.}
\label{tab:LVMablationstudy}
\end{table}

\begin{figure*}[!t]
    \centering
    \includegraphics[scale=0.6]{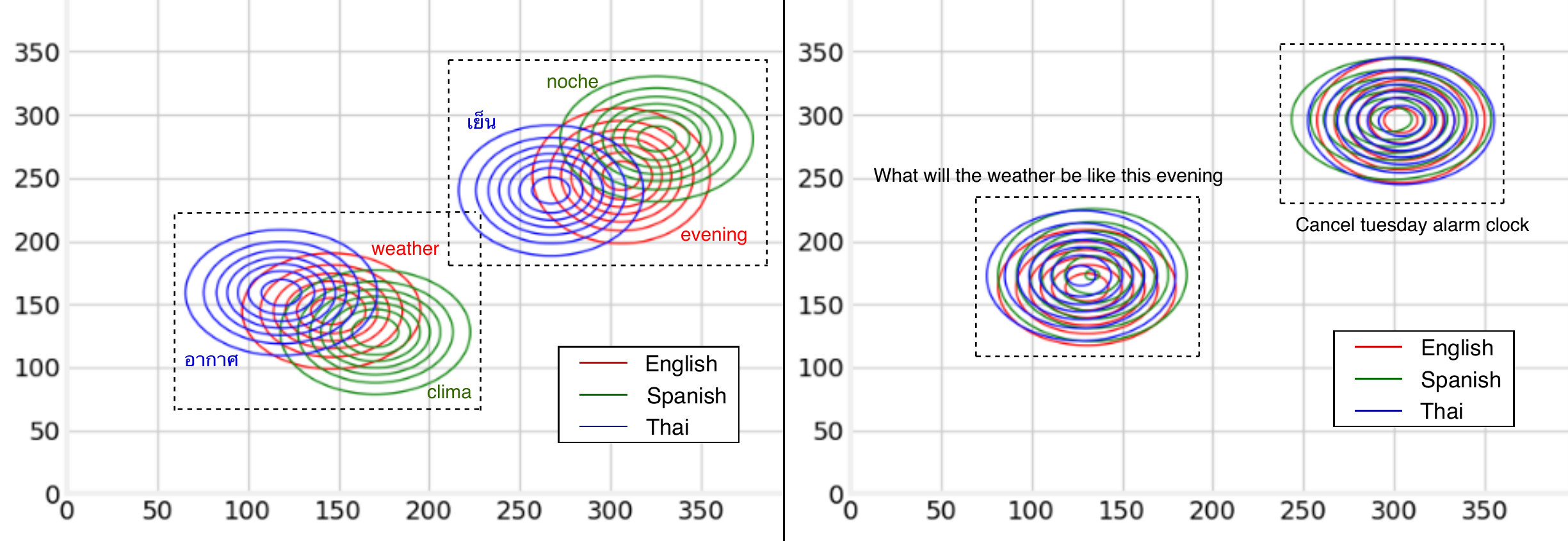}
    \caption{Visualization of latent variables on words \textbf{(left)} and sentences \textbf{(right)}. \textbf{Left:} We choose ``weather-clima-
    \foreignlanguage{thaicjk}{อากาศ}'' 
    and ``evening-noche-
    \foreignlanguage{thaicjk}{เย็น''}
    from parallel sentences. \textit{English: ``What will the weather be like this evening''}, \textit{Spanish: ``Cómo será el clima esta noche''}, 
    \textit{Thai: }``\foreignlanguage{thaicjk}{ตอน เย็น นี้ อากาศ จะ เป็น อย่างไร''}.
    \textbf{Right:} We choose two English sentences and show their distributions and those of the corresponding Spanish and Thai translations.}
    \label{fig:visual}
\end{figure*}

\subsection{Evaluation}
We implement and evaluate the following models:

\paragraph{\textbf{Zero-shot SLU}} \citet{upadhyay2018almost} used cross-lingual embeddings \cite{bojanowski2017enriching} to do zero-shot transfer learning. 
\paragraph{\textbf{Conditional Random Fields (CRF)}} We reproduce the baseline model in \citet{schuster2019cross}, and also add embedding noise, cross-lingual refinement, and delexicalization.
\paragraph{\textbf{Latent Variable Model (LVM) - Ours}} We replace the CRF module with latent variables and also apply it to intent prediction.

\medskip

\noindent Besides, we directly compare with the baseline models illustrated in~\citet{schuster2019cross}:

\paragraph{\textbf{Multi. CoVe w/ auto}} They combined Multilingual CoVe~\cite{yu2018multilingual} with an auto-encoder objective and then used the trained encoder with the CRF model. 
\paragraph{\textbf{Translate Train}} They trained a supervised machine translation system to translate English data into the target language and then trained the CRF model on this translated dataset.

\section{Results \& Discussion}

From Table~\ref{tab:result}, in general, LVM outperforms CRF models. This is because for semantically same words (e.g., weather and clima) LVM considers such close enough points as the same distribution, but CRF is more likely to classify them differently. This can be shown very clearly from Figure~\ref{fig:visual}, in which the latent variables demonstrate similar distributions for semantically similar sentences and words. In addition, we can see that adding only Gaussian noise to the Vanilla BiLSTM improves our prediction performance significantly, which implies that the robustness of our model towards the noisy signals which come from the target embedding inputs. 

Furthermore, it is clearly visible that cross-lingual embeddings refinement is more effective in Spanish than Thai. This is attributed to the quality of alignments in the two languages. Spanish is much more lexically and grammatically similar to English than Thai, so word-level embedding refinement is reasonably good. Jointly incorporating all three methods (Gaussian noise injection, cross-lingual embeddings refinement, and delexicalization) further reduces the noise in the inputs as well as makes the model more robust to noise, which help LVM to more easily approximate the distribution.

Finally, in Table~\ref{tab:LVMablationstudy}, we ablate the usage of LVM to see whether the boost of performance comes simply from the increase of parameter size. By removing or replacing LVM with MLP, we can see the clear performance gains by using LVM.

\section{Conclusion}
In this paper, we propose a transferable latent variable that focuses on improving the zero-shot cross-lingual adaptation of natural language understanding task to low-resource languages. We show that a combination of 1) cross-lingual embeddings refinement, 2) Gaussian noise injection, and 3) latent variables are effective in coping with the variance of semantically similar sentences across different languages, and the visualizations of the latent variables confirm such. We leverage very few resources (i.e., 11 seed word pairs) and achieve state-of-the-art performance for English-to-Spanish and English-to-Thai in the zero-shot cross-lingual scenario.

\section*{Acknowledgments}
This work has been partially funded by ITF/319/16FP and MRP/055/18 of the Innovation Technology Commission, the Hong Kong SAR Government.




\bibliography{emnlp-ijcnlp-2019}
\bibliographystyle{acl_natbib}

\end{document}